\def\year{2018}\relax
\documentclass[letterpaper]{article} 
\usepackage{aaai18}  
\usepackage{times}  
\usepackage{helvet}  
\usepackage{courier}  
\usepackage{url}  
\usepackage{graphicx}  
\usepackage{multirow}
\usepackage{multicol}
\usepackage{array}
\usepackage{graphicx}
\usepackage{footnote}
\usepackage{amsmath, bm}
\usepackage{color}
\usepackage{xcolor}
\usepackage{xspace}
\usepackage[english]{babel}
\usepackage{subfig}
\usepackage{booktabs}
\usepackage{tabularx}
\usepackage{mathtools}
\usepackage{color, colortbl}

\definecolor{Gray}{gray}{0.9}
\DeclarePairedDelimiter{\ceil}{\lceil}{\rceil}

\newcommand{\DS}{\texttt{DS}\xspace}
\newcommand{\KB}{\texttt{KB}\xspace}
\newcommand{\KBs}{\texttt{KBs}\xspace}

\newcommand{\FB}{\texttt{FB}\xspace}
\newcommand{\CVT}{\texttt{CVT}\xspace}

\begin{document}
\author{Ying Zeng $^1$ \and Yansong Feng $^{1,*}$ \and Rong Ma $^1$ \and Zheng Wang $^2$ \\ 
	{\bf \large Rui Yan $^1$ \and Chongde Shi $^3$ \and Dongyan Zhao $^1$}\\
 	$^1$ Institute of Computer Science and Technology, Peking University, P.R. China,\\
 	$^2$ School of Computing and Communications, Lancaster University, UK,\\
 	$^3$ Institute of Scientific and Technical Information of China\\
	\{ying.zeng, fengyansong, marongcs, ruiyan, zhaody\}@pku.edu.cn\quad z.wang@lancaster.ac.uk\quad shcd@istic.ac.cn
}

\title{Scale Up Event Extraction Learning via Automatic Training Data Generation}

\maketitle
\begin{abstract}
	The task of event extraction has long been investigated in a supervised learning paradigm, which is bound by the number and the quality of the training instances. Existing training data must be manually generated through a combination of expert domain knowledge and extensive human involvement. However, due to drastic efforts required in annotating text, the resultant datasets are usually small, which severally affects the quality of the learned model, making it hard to generalize.
	Our work develops an automatic approach for generating training data for event extraction. Our approach allows us to scale up event extraction training instances from thousands to hundreds of thousands, and it does this at a much lower cost than a manual approach. We achieve this by employing distant supervision to automatically create event annotations from unlabelled text using existing structured knowledge bases or tables. We then develop a neural network model with post inference to transfer the knowledge extracted from structured knowledge bases to automatically annotate typed events with corresponding  arguments in text. We evaluate our approach by using the knowledge extracted from Freebase to label texts from Wikipedia articles. Experimental results show that our approach can generate a large number of high-quality training instances. We show that this large volume of training data not only leads to a better event extractor, but also allows us to detect multiple typed events.
\end{abstract}

\section{Introduction}
Event extraction is a key enabling technique for many natural language processing (NLP) tasks. The goal of event extraction is to detect,
from the text, the occurrence of events with specific types, and to extract arguments (i.e. typed participants or attributes) that are
associated with an event. Current event extractor systems are typically built through applying supervised learning to learn over labelled
datasets. This means that the performance of the learned model is bound by the quality and coverage of the training datasets.

To generate training data for event extraction, existing approaches all require manually identifying -- if an event occurs and of what type
-- by examining the event trigger\footnote{A trigger is the word or phrase that clearly expresses the occurrence of an event. E.g.,
	\textit{ex} in \textit{ex-husband} triggers a \emph{divorce} event.} and arguments from each individual training instance. This process
requires the involvement of linguists to design annotation templates and rules for a set of predefined event types, and the employment of
annotators to manually label if an individual instance belongs to a predefined event type.

To determine an event type, existing approaches  all require \emph{explicitly} identifying event triggers from text and assign them with predefined types, 
thus relies on human to annotate training data. This makes the quality and the number of the generated instances dependent on the skill and available time of the
annotators~\cite{aguilar2014comparison,song2015light}. To scale up event extractor training, we therefore must take human annotators out
from the loop of data labeling.

This work develops a novel approach to automatically label training instances for event extraction. Our key insight was that while event
triggers are useful, they do not always need to be explicitly captured. Structured knowledge bases (\KBs) such as Freebase already provide
rich information of event arguments, organizing as structured tables, to enable us to automatically infer the event type. For example,
sentence ``\textit{In 2002, WorldCom made its filing for Chapter 11 in New York.}'' describes a \emph{bankruptcy} event; but this event
does not have to be identified through a \emph{bankrupt} trigger (and in fact such a trigger is missing in this sentence) -- in this case,
several key arguments together also imply such an event.

If we can find ways to exploit structured tables or lists, a single entry can then be used to label many instances without
human annotations. Such an approach is known as distant supervision (\DS). Recent
studies~\cite{mintz2009distant,zeng2015distant} have demonstrated its effectiveness in various NLP tasks. The central idea of \DS is to use
the knowledge extracted from a known \KB to \emph{distantly supervise} the process of training data generation.

We observe that the structured tables of Freebase and Wikipedia can be useful for inferring event types. We then design heuristics to
identify what are the most important properties, or \emph{key arguments},  of a table entry in determining the occurrence of an event. We
show that it is possible to completely forgo explicit trigger information and entirely rely on key arguments to determine event types.
Using key arguments, we can now develop an \emph{automatic} approach to generate training data for event extraction. Under this new \DS
paradigm, we further propose a novel event extractor architecture based on neural networks and linear integer programming. One of the key
innovations of our approach is that the model does not rely on explicit triggers. Instead, it uses a set of key arguments to characterize
an event type. As a result, it eliminates the need to explicitly identify event triggers, a process that requires heavy human involvement.

We evaluate our approach by applying it to sentence-level event extraction. We have conducted intensive experiments on multiple datasets
generated using various knowledge resources. Our experimental results confirm that key arguments are often sufficient enough for infering
event types. 
Using the structured tables from FreeBase and Wikipedia, we are able to automatically generate a large number of training
instances -- resulting in a training dataset that is 14x greater than the widely used ACE Challenge dataset\cite{doddington2004automatic}, and our dataset was automatically constructed within hours instead of costing years of linguists and annotators' time. We show that the quality of the automatically
generated data is comparable to ones that would be manually constructed by human experts. Using the larger volume of training data, in
combination of our novel event extractor architecture, we can not only learn a highly effective event extraction system, but also unlock
the potential of multiple typed event detection -- a feature that few of the existing event extraction methods supports, but is much needed.

\section{Motivation}
\begin{figure}
	\centering
	\subfloat[][An example sentence from Wikipedia]{\includegraphics[width=0.47\textwidth]{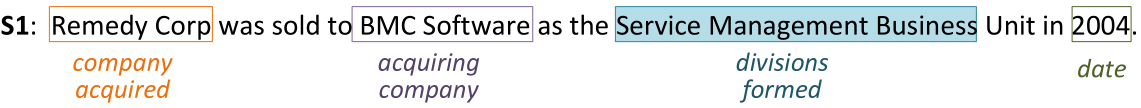}}
	\\
	\subfloat[][Entry of \texttt{business.acquisition} in Freebase]{
		\scriptsize
		\begin{tabular}{lp{35pt}p{35pt}lp{60pt}}
			\toprule
			id & company\_ acquired & acquiring\_ company & date & divisions\_ formed\\
			\midrule
		 m.07bh4j7 & Remedy Corp & BMC \quad Software & 2004 & Service Management Business Unit\\
			\bottomrule
		\end{tabular}
	}
	\caption{The event type of the sentence shown in (a) can be automatically inferred using the structured table given by Freebase in (b).}
	\label{fig:example}
\end{figure}

As a motivation example, consider the sentence 
in Figure~\ref{fig:example}(a). To detect the event,
traditional approaches require first identifying the trigger word, \emph{sold}, then assigning it with a type of \texttt{business.acquisition}.

For this particular sentence, we argue that identifying the trigger is not necessary for determining the event type.
Figure~\ref{fig:example} (b) shows a Compound Value Type (\CVT) entry from Freebase (\FB)~\cite{bollacker2008freebase}. Here, a \CVT
organizes complex structured data with multiple \emph{properties} in a table\footnote{Therefore, we also use the term ``argument" to refer
	to a \CVT property for the rest of paper.}. Using this \CVT schema, one can easily map the three arguments of sentence S1, \emph{Remedy
	Corp}, \emph{BMC Software}, and \emph{2004}, respectively to their properties, \texttt{company\_acquired}, \texttt{acquiring\_company}, and
\texttt{date}. Here each property essentially describes the role of an argument in the sentence; and in combination, they define a
\texttt{business.acquisition} event.

This example shows that \CVT information can be used to infer event type without explicit trigger information. If we can utilize such \CVT
information, we can then label sentences without needing to explicitly identifying any triggers. As a result, we can free annotators from
the labour-intensive manual process. In this work, we develop a simple, yet effective technique to automatically generate training data by
exploiting the prior \CVT knowledge.

\section{Training Data Generation}
Our approach exploits structural information like \FB \CVT tables to  automatically annotate event mentions\footnote{An event mention is a phrase or sentence within which an event is described, including its type and arguments.}, and generate training data to learn an
event extractor. We use the arguments of a \CVT table entry to infer what event a sentence is likely to express. A \CVT table entry can
have multiple arguments, but not all of them are useful in determining the occurrence of an event. For instance, the
\texttt{divisions\_formed} argument in Figure~\ref{fig:example}(b) is not as important as the other three arguments when determining if a
sentence expresses a \texttt{business.acquisition} event.

Our first step for training data generation is to identify the key arguments from a \CVT table entry. A \textbf{key argument} is an
argument that plays an important role in one event, which helps to distinguish with other events. If a sentence contains all key arguments
of an entry in an event table (e.g. a \CVT table), it is likely to express the event presented by the table entry. If a sentence is
labelled as an event mention of a \CVT event, we also record the words or phrases that match the entry's properties as the involved
arguments, with the roles specified by their corresponding property names.

\subsection{Determining Key Arguments}
We use the following formula to calculate the importance score, $I_{cvt, arg}$, of an argument \emph{arg} (e.g., date) to its event type
\emph{cvt} (e.g., \texttt{business.acquisition}):

\begin{equation}
I_{cvt, arg} = log \frac{count(cvt, arg)}{count(cvt) \times count(arg)}
\end{equation}

where $count(cvt)$ is the number of instances of type $cvt$ within a \CVT table, $count(arg)$ is the number of times $arg$ appearing in all
\CVT types within a \CVT table, and $count(cvt, arg)$ is the number of $cvt$ instances that contain $arg$ across all \CVT tables.

\begin{figure}
	\centering
	\includegraphics[width=.47\textwidth]{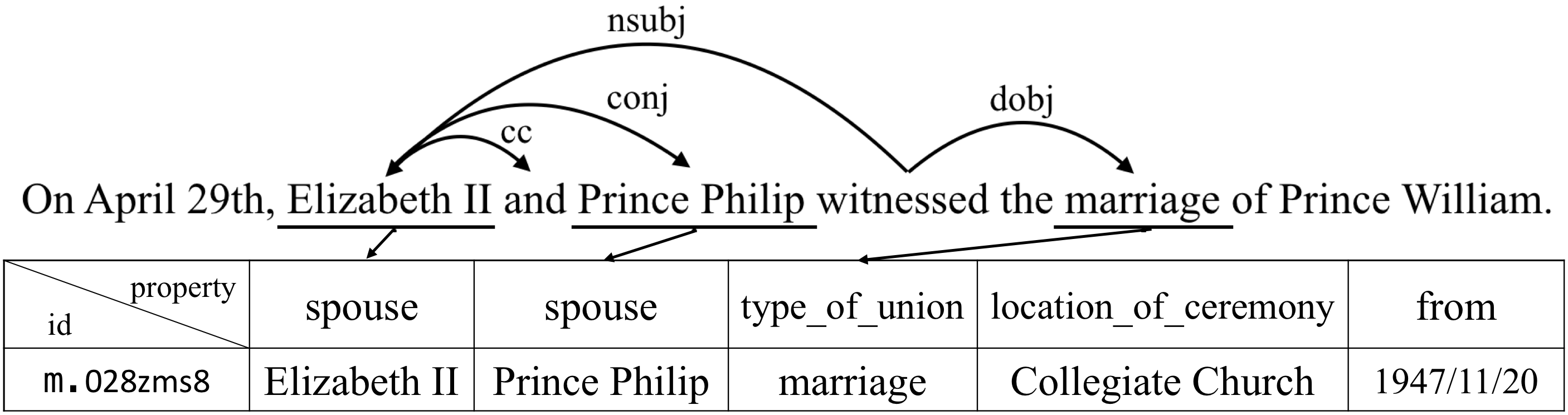}
	\caption{The dependency tree of S4, which partially matches a \CVT entry of \emph{people.marriage} in \FB. \label{fig:2}}
\end{figure}

Our strategy for selecting key arguments of a given event type is described as follows:

\begin{description}
	
	\item [P1] For a \CVT table with $n$ arguments, we first calculate the importance score of each argument. We then consider the top half
	$\ceil[\big]{n/2}$ (rounding up) arguments that have the highest importance scores as key arguments.
	
	\item [P2] We find that time-related arguments are useful in determining the event type, so we always include a time-related argument\footnote{If there are multiple time-related arguments, we select the one with highest importance score.}
	(such as date) in the key argument set.
	
	\item [P3] We also remove sentences from the generated dataset in which the dependency distances between any two key arguments are
	greater than 2. The distance between two arguments is the minimal number of hops it takes from one argument to the other on the dependency parse tree (see also Figure~\ref{fig:2}).
	
\end{description}

Using this strategy, the first three arguments of the \CVT entry are considered to be key arguments for event type
\texttt{business.acquisition}.

\paragraph{Selection Criteria: }
To determine how many arguments should be chosen in P1, we have conducted a series of evaluations on the quantity and quality of the datasets using different policies. We found that choosing the top half arguments gives the best accuracy for event labeling.

We use the following three example sentences from Wikipedia to explain P2 and P3 described above. 
\begin{quote}
	\textbf{S2}: \underline{\emph{Microsoft}} spent \$6.3 billion buying online display advertising company \underline{\emph{aQuantive}} in
	\underline{\emph{2007}}.
\end{quote}
\begin{quote}
	\textbf{S3}: Microsoft hopes aQuantive's Brian McAndrews can outfox Google.
\end{quote}
\begin{quote}
	\textbf{S4}: On April 29th, Elizabeth II and Prince Philip witnessed the marriage of Prince William.
\end{quote}

\begin{table}
	\scriptsize
	\centering
	\begin{tabular}{lp{40pt}p{40pt}lp{50pt}<{\centering}}
		\toprule
		 id &  company\_ acquired & acquiring\_ company &  date & divisions\_formed\\
		\midrule
		m.05nb3y7 & aQuantive & Microsoft & 2007 & N/A\\
		\bottomrule
	\end{tabular}
	\caption{A \texttt{business.acquisition} \CVT entry in \FB. \label{tbl:bs}}
\end{table}

Although time-related arguments are often missing in the currently imperfect \KBs, they are crucial to identify the actual occurrence of an
event. As an example, suppose we want to use the \CVT entry shown in Table~\ref{tbl:bs} to determine whether sentences S2 and S3 are
\texttt{business.acquistion} events. While this strategy works for S2, it gives a false prediction for S3. If we add the time-related
argument (i.e., date: 2007), we can then correctly label both sentences.
Therefore, we always consider the most important time-related argument as key arguments (P2) when they are available.

Finally, P3 is based on our intuitions that two arguments involved in the same event mention are likely to be closer within the
syntactic structure.
As shown in  Figure~\ref{fig:2}, although both \emph{Prince Philip} and \emph{marriage} can be matched as key arguments in a  \texttt{people.marriage}
entry, but with a distance of 3 (i.e., far from each other under our criterion) on the dependency tree, thus S4 will be labeled as
negative.

\subsection{Data Generation}
To generate training data, we follow a number of steps.

Our approach takes in existing structured tables or lists that are organized in a way similar to the \FB \CVT tables. The structured tables
can be obtained from an existing knowledge base, created by experts, or generated through a combination of both approaches. For each event
type, we determine the key arguments for each entry within that type used the key argument selection strategy described above. This step
produces a set of rules to be used for data labeling, where each rule contains the event type, key arguments and non-key arguments given by
a structured table entry. We also use alias information (such as Wikipedia redirect) to match two arguments that have different literal
names but refer to the same entity (e.g. Microsoft and MS).

Next, we label each individual sentence from the target dataset. The labeling process is straightforward. We enumerate all rules from the
generate rule set, and check if the target sentence contains all the key arguments specified by a rule. We regard a sentence as a
\emph{positive} sample if it contains all the key arguments of a rule, or \emph{negative} otherwise. For instances, S1 in Figure~\ref
{fig:example}(a) and S2 (with its arguments in italics and underlined) are positive examples, while S3 and S4 are negative.

Because 68\%  of  arguments in our dataset consist of more than one word, we formulate the training in a sequence labeling paradigm rather
than word-level classifications. We tag each word of the sentence using the standard begin-inside-outside (\texttt{BIO})
scheme, where each token is labeled as \texttt{B-role} if it is the beginning of an event argument with its role \texttt{role}, or \texttt{I-role} if it is inside a \texttt{role}, or \texttt{O} otherwise. We call this a \textbf{labeling sequence}.

\subsection{Limitations}
Our approach relies on structured tables or lists to automatically label text. The table can be obtained from an existing \KB or
hand-crafted by experts. We stress that providing a table incurs much less overhead than manually tagging each training sentence, as a
single table entry can be automatically applied to an unbounded number of sentences.

Our implementation may benefit from pronoun resolution  and entity coreference, which is complementary to our method, and may improve our recall.
While this work targets the sentence level, we believe it is generally applicable. There are methods like event coreference resolution\cite{liao2010using,berant2014modeling},  can be used to extend our approach to document-level event extraction.

\section{Event Extraction}
\begin{figure}[t!]
	\centering
	\includegraphics[width=0.47\textwidth]{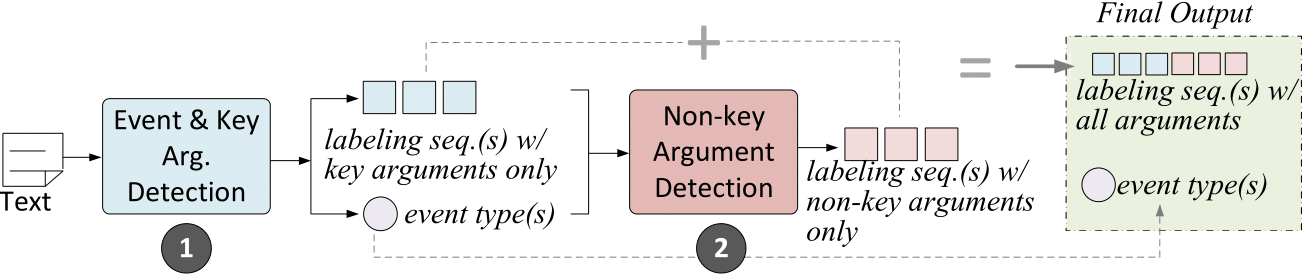}
	\caption{Our 2-stage event extraction pipeline.}\label{fig:model}
\end{figure}

Unlike prior works, our approach does not rely on explicit trigger identification. Instead, it uses key arguments to detect the occurrence of
an event. We solve the problem in a 2-stage pipeline, depicted in Figure~\ref{fig:model}, which takes raw text as input and outputs labeling sequence(s) and event type(s)-- if any event is detected.

The first stage identifies the key arguments in a sentence. If a sentence contains \emph{\textbf{all} key arguments} of a specific event type, it will be considered to imply an event mention of this specified type.
Since at this stage we do not concern non-key arguments, all non-key argument tokens are tagged as \texttt{O}. Furthermore, multiple labeling sequences may be produced by this stage, each corresponds to an event type. The second stage takes in the outputs of the first stage, and detects all the non-key arguments.

\subsection{Stage 1: Key Argument and Event Detection \label{evede}}
The model used in stage 1 consists of a Bidirectional Long Short-Term Memory (BLSTM) network with a conditional random field \cite{lafferty2001conditional}  (CRF) layer
and an Integer Linear Programming (ILP) based post inference. The BLSTM-CRF layer finds the optimal labeling sequence which will then be
post-processed by an ILP solver. Using the labeling sequence, an event detector then checks if the sentence mentions a specific event.

\paragraph{BLSTM}
At each time $t$, a forward LSTM layer takes $\textbf{x}_t$ as input and computes the output vector $\overrightarrow{\textbf{h}}_t$ of the
past context, while a backward LSTM layer reads the same sentence in reverse and outputs $\overleftarrow{\textbf{h}}_t$ given the future
context. We concatenate these two vectors to form the output vector of a BLSTM, which is fed into a softmax layer to estimate a probability
distribution over all possible labels.

\paragraph{CRF}
Choosing the best label for each word individually according to the BLSTM ignores the dependencies between labels, thus cannot guarantee the best sequence. 
Therefore, we introduce a CRF layer over the BLSTM output.

We consider $\textbf{P}$ to be a matrix of confidence scores output by BLSTM, and the element $\textbf{P}_{i,j}$ of the matrix denotes the probability of the label $j$ for the $i$-th word in a sentence. The CRF layer takes a transition matrix $\textbf{A}$ as parameter, where $\textbf{A}_{i,j}$ represents the score of a transition from label $i$ to label $j$. The score of a sentence $\bm{w}$ along with a path of labels $\bm{y} = \{y_1, y_2, \ldots, y_n\}$ is measured by the sum of BLSTM outputs and transition scores:
\begin{equation}
score(\bm{w}, \bm{y}) = \sum\limits_{i=0}^n\textbf{P}_{i, y_i} + \sum\limits_{i=1}^n\textbf{A}_{y_i, y_{i+1}}
\end{equation}
During test, given a sentence $\bm{w}$, we adopt the Viterbi algorithm~\cite{rabiner1989tutorial} to find the optimal label sequence with the maximum score among all possible label sequences.

\paragraph{ILP-based Post Inference}
The output sequences of BLSTM-CRF do not necessarily satisfy the structural constraints of event extraction.
We thus propose to apply ILP to further globally optimize the BLSTM-CRF output  to produce the best label sequence. Formally, let
$\mathcal{L}$ be the set of possible argument labels. For each word $w_i$ in the sentence $\bm{w}$ and a pair of labels $ \langle l, l'
\rangle \in \mathcal{L} \times \mathcal{L}$, we create a binary variable ${v_{i,l,l'} \in \{0, 1\}}$, denoting whether or not the $i$-th
word $w_i$ is tagged as label $l$ and its following word $w_{i+1}$ is tagged as label $l'$ at the same time. The objective of ILP is to
maximize the overall score of the variables as:
\begin{displaymath}
\sum\nolimits_{i, l, l'}v_{i,l,l'} * (\textbf{P}_{i,l}+\textbf{A}_{l,l'}) .
\end{displaymath}
where we consider the following four constraints:

\textbf{C1}: Each word should be and only be annotated with one label, i.e.:
\begin{equation}
\sum\nolimits_{l,l'}v_{i,l,l'}=1
\end{equation}

\textbf{C2}: If the value of $v_{i,l,l'}$ is $1$, then there has to be a label $l^*$ that will make $v_{i+1,l',l^*}$ equal to $1$, i.e.:
\begin{equation}
v_{i,l,l'} = \sum\nolimits_{l^*}v_{i+1,l',l^*}
\end{equation}

\textbf{C3}: If the current label is \texttt{I-arg}, then its previous label must be \texttt{B-arg} or \texttt{I-arg}, i.e.:
\begin{equation}
v_{i,\texttt{I-arg},l'} = v_{i-1,\texttt{B-arg},\texttt{I-arg}} + v_{i-1, \texttt{I-arg}, \texttt{I-arg}}
\end{equation}

\textbf{C4}: For a specific event type, all its key arguments should co-occur in the sentence, or none of them appears in the resulting sequence. For any pair of key arguments $arg_1$ and $arg_2$ with respect to the same event type, the variables related to them are subject to:
\begin{equation}
\sum\nolimits_{i,l'}{v_{i,\texttt{B-arg}_1,l'}} \leq n * \sum\nolimits_{j,l^*}{v_{j,\texttt{B-arg}_2,l^*}}
\end{equation}
where $n$ is the length of the sentence.

\paragraph{Event Detection}
This step simply checks if the input sentence contains all the key arguments of a specific event type by
examining the labeling sequence. 

\paragraph{Multiple Typed Events}
There are scenarios where one sentence expresses multiple events which share some key arguments, but most current event extractors only map a sentence to one event.
For example, in S5, \emph{Kevin Spacey} is the \texttt{actor} of a \texttt{film\_performance} event and a \texttt{tv\_appearance} event triggered by the same word \emph{stared}.
\begin{quote}
	\textbf{S5}: \underline{\emph{Kevin Spacey}} stared as \underline{\emph{Frank Underwood}} in the Netflix series \underline{\emph{House of Cards}}, and later as \underline{\emph{Tom Brand}} in \underline{\emph{Nine Lives}}.
\end{quote}
One of the advantages of our approach is that it can be easily extended to support multiple typed events.
To do so, we allow our ILP solver to output multiple optimal sequences. Specifically, after our model outputs the best sequence $\bm{s}^t$
at time $t$, we remove the previously best solutions
$\{\bm{s}^1, \ldots, \bm{s}^{t}\}$ from the solution space, and re-run our solver to obtain the next optimal sequences $\bm{s}^{t+1}$.
We repeat the optimization procedure until the difference between the scores of $\bm{s}^1$ and $\bm{s}^T$ is greater
than a threshold $\lambda$, and consider all solutions $\{\bm{s}^1, \bm{s}^2, \ldots, \bm{s}^{T-1}\}$ as the optimal label sequences.
We use Gurobi~\cite{gurobi} as our ILP solver and set $\lambda=0.5 \times n$, which averagely produce~1.07 optimal sequences for each sentence.

\subsection{Stage 2: Non-key Argument Detection}
After event detection, a sentence will be classified into one or more event types, and labeled with the corresponding key arguments.
We next adopt the same BLSTM-CRF architecture to detect the remaining non-key arguments, where we encode the key-argument label (from the first stage) of each word into a key-argument feature vector through a look-up table, and concatenate it with the original word embedding as the input to a new BLSTM-CRF. Note that we do not need post inference here, because there is no structural constraints between non-key arguments.

\section{Experiments}
Our experiments are designed to answer: (1) whether it is possible to automatically collect training data for event extraction, (2) whether
extractors trained on such data can detect events of interest and identify their corresponding arguments, and (3) whether our solution can
work with other knowledge resources for more types of event.

\subsection{Dataset Evaluation}\label{sec:evalhypo}
We start by comparing key argument selection strategies:
(1) \emph{ALL}: uses all arguments as key arguments; (2) \emph{IMP}: uses the top half arguments with highest importance scores as key
arguments; (3) \emph{IMP\&TIME}: includes a time-related argument together with  the arguments selected by \emph{IMP}; and (4) \emph{DIS}: eliminate sentences where the dependency distances between any two key arguments are greater than 2.
We use the above methods to collect datasets using Freebase and the English Wikipedia dump of 2016-11-20, by randomly selecting 100
sentences from each dataset, and ask two annotators to decide if each sentence implies a given event type.

As shown in Table~\ref{tab:3}, it is not surprising that \emph{ALL}, as the most strict, guarantees the quality of the collected data, but only contributes 203 sentences covering 9 event types, which is far from sufficient for further applications. \emph{IMP} relaxes \emph{ALL} by allowing the absence of non-key arguments, which expands the resulting dataset, but introduces more noise.
We can also see that the dependency constraint (DIS) improves the data quality (\emph{IMP}+\emph{DIS}).
Compared with \emph{IMP}, the significant quality improvement by \emph{IMP\&TIME} proves that time-related arguments within CVT schemas are critical to imply an event occurrence. Among all strategies, the dataset by \emph{IMP\&TIME}+\emph{DIS}  achieves the best quality, while still accounting for 46735 sentences with 50109 events, almost 10 times more than the ACE dataset, showing that it is feasible to automatically collect quality training data for event extraction without either human-designed event schemas or \textbf{extra} human annotations.

Our final dataset, \textbf{FBWiki}, using \emph{IMP\&TIME}+\emph{DIS} , contains 46,735 positive sentences and 79,536 negative ones\footnote{Besides trivial negative samples that have no matched arguments, we randomly sample 34,837 negative instances that contain only part of key arguments, and 21,866 sentences whose key arguments violate the dependency constraint.} a random split of 101,019 for training and 25,252 for testing. There are on average 4.8 arguments per event, and in total, 5.5\% instances labeled with more than two types of events.

\begin{table}[t!]
	\scriptsize
	\centering
	\begin{tabular}{lccc}
		\toprule
		\textbf{Strategy} & \textbf{Sentences} & \textbf{Type} & \textbf{Positive Percentage (\%)} \\
		\midrule
		\rowcolor{Gray}\emph{ALL} & 203 & 9 & 98\% \\
		\emph{IMP} & 318K & 24 & 22\% \\
		\rowcolor{Gray} \emph{IMP}+\emph{DIS} & 170K & 24 & 37\% \\
		\emph{IMP\&TIME} & 112K & 24 & 83\% \\
		\rowcolor{Gray} \emph{IMP\&TIME}+\emph{DIS} & 46K & 24 & 91\% \\
		\bottomrule
	\end{tabular}
	\caption{Statistics of the datasets built with different strategies.
		\textit{Type} the number of different CVT types found.
		\label{tab:3}}
\end{table}

\noindent \textbf{\emph{Trigger Inference}: \mbox{ }} To further explore the relationship between key arguments and triggers, we regard the
least common ancestor of all key arguments in the dependency tree as a trigger candidate. As listed in Table~\ref{freqTriggers}, these
candidates share similar meanings and are highly informative to the underlying event types, showing that our key arguments with necessary
constraints can play the same role with explicit triggers in identifying an event.

\begin{table}
	\scriptsize
	\centering
	\begin{tabular}{lll}
		\toprule
		\textbf{Event types} & \textbf{Trigger candidates} & \textbf{Percentage} \\
		\midrule
		\rowcolor{Gray}film\_performance & play, appear, star, cast, portray & 0.72 \\
		award\_honor & win, receive, award, share, earn & 0.91 \\
		\rowcolor{Gray}education & graduate, receive, attend, obtain, study & 0.83 \\
		acquisition & acquire, purchase, buy, merge, sell & 0.81 \\
		\rowcolor{Gray}employ.tenure & be, join, become, serve, appoint & 0.79 \\
		\bottomrule
	\end{tabular}
	\caption{Top 5 most frequent trigger candidates and their proportions over all positive instances within each type.}
	\label{freqTriggers}
\end{table}

\noindent \textbf{\emph{On ACE}: \mbox{ }} We also test our strategy on the ACE dataset. We first collect all annotated events,
without triggers, as the knowledge base to compute the importance values for all arguments, and select the key arguments for each ACE event
type accordingly. We follow \emph{IMP\&TIME}+\emph{DIS} to examine every sentence whether it can be selected as an annotated instance
within the ACE event types. Eventually, we correctly obtain 3,448 sentences as positive instances, covering 64.7\% of the original ACE
dataset.  We find that the main reason for the missing 35.3\% is that many arguments in the ACE dataset are pronouns, where our strategy is
currently unable to treat pronouns as key arguments. However, if a high-precision coreference resolution tool is available to preprocess
the document, our solution would be able to automatically label more instances.

\subsection{Extraction Setup}\label{sec:evalevent}
Next, we evaluate our event extractor on FBWiki with \emph{precision} (P), \emph{recall} (R), and \emph{F-measure} (F) for each subtask. These metrics are computed according to the following standards of correctness. For \emph{event classification}, an event is correctly classified if its reference sentence contains
\textbf{all key arguments} of this event type. For \emph{key argument detection}, an event is correctly detected if its type and all of its
key arguments match a reference event within the same sentence. For \emph{all argument detection}, an event is correctly extracted if its
type and all of its arguments match a reference event within the same sentence.

\noindent \textbf{\emph{Training}:  \mbox{ }} All hyper-parameters are tuned on a development split in the training set. During event detection, we
set the size of word embeddings to 200, the size of LSTM layer to 100. In argument detection, we use the same size of word embedding, while
the size of LSTM layer is 150, and the size of key argument embedding is 50. Word embeddings are pre-trained using skip-gram
word2vec~\cite{mikolov2013distributed} on English Wikipedia and fine tuned during training. We apply dropout (0.5) on both input and output
layers.

\noindent \textbf{\emph{Baselines}:  \mbox{ }} We compare our proposed model with three baselines.
The first is a  BLSTM model that takes word embeddings as input, and outputs the label for each word with the maximum probability. 
For feature-based methods, we apply CRF (using the CRF++ toolkit~\cite{kudo2005crf++} ) and Maximum Entropy \cite{berger1996maximum} (Le
Zhang's MaxEnt toolkit) to explore a variety of elaborate features, according to the state-of-art feature-based ACE event
extractors~\cite{li2013joint}. Note that after key argument detection, we add the resulting label of each word as a supplementary feature
to detect non-key arguments.

\subsection{Compare with Automatic Annotations}
Firstly, we compare the model output against the automatically obtained event annotations.
As shown in Table~\ref{tab:1}, feature-based models perform worst in both event classification and argument detection.
One of the main reasons is the absence of explicit trigger annotations in our dataset, which makes it impossible to include trigger-related features, e.g., trigger-related dependency and position features.
Although traditional models can achieve higher precisions, they only identify a limited number of events, resulting in low recalls.
Neural-network methods perform much better than feature-based models, especially in recall, since they can make better use of word semantic features. Specifically, BLSTM can capture longer dependencies and richer contextual information, instead of neighbouring word features only.
The CRF layer brings an averagely 2\% improvement in all metrics, and by adding the ILP-based post inference, our full model, BLSTM-CRF-ILP$_{multi}$, achieves the best performance among all models.

\begin{table*}[!t]
	\centering
	\scriptsize
	\begin{tabular}{|l|p{1.cm}<{\centering}|p{1.cm}<{\centering}|p{1.cm}<{\centering}|p{1.cm}<{\centering}|p{1.cm}<{\centering}|p{1.cm}<{\centering}|p{1.cm}<{\centering}|p{1.cm}<{\centering}|p{1.cm}<{\centering}|} \hline
		\multirow{2}{*}{\textbf{Model}} & \multicolumn{3}{c|}{\textbf{Event Type}} & \multicolumn{3}{c|}{\textbf{Key Argument Detection}} &
		\multicolumn{3}{c|}{\textbf{All Argument Detection}} \\ \cline{2-10}
		& \textbf{P} & \textbf{R} & \textbf{F} & \textbf{P} & \textbf{R} & \textbf{F} & \textbf{P} & \textbf{R} & \textbf{F }\\ \hline
		\rowcolor{Gray} CRF & 88.9 & 11.0 & 19.6 & 36.1 & 4.47 & 7.96 & 19.9 & 3.06 & 5.30  \\ \hline
		MaxEnt & \textbf{95.2} & 12.4 & 21.9 & 41.6 & 5.40 & 9.56 & 22.5 & 3.40 & 5.91 \\ \hline
		\rowcolor{Gray} BLSTM & 89.8 & 63.0 & 74.1 & \textbf{64.9} & 45.5 & 53.5 & 42.9 & 27.7 & 33.7  \\ \hline \hline
		BLSTM-CRF & 86.4 & 67.4 & 75.7 & 63.6 & 49.6 & 55.8 & \textbf{44.4} & 31.0 & 36.5  \\ \hline
		\rowcolor{Gray} BLSTM-CRF-ILP$_{1}$ & 84.4 & 74.1 & 78.9 & 62.3 & 53.8 & 57.3 & 42.7 & 33.8 & 37.7 \\ \hline
		BLSTM-CRF-ILP$_{multi}$ & 85.3 & \textbf{79.9} & \textbf{82.5} & 60.4 & \textbf{55.3} & \textbf{57.7} & 41.9 & \textbf{34.6} & \textbf{37.9} \\ \hline
	\end{tabular}
	\caption{System performance when compared against automatic annotations (\%).  \label{tab:1}}
\end{table*}

It is not surprising that  every model with a CRF layer over its BLSTM layer is superior to the one with a BLSTM layer only. Compared with vanilla BLSTM, BLSTM-CRF achieves higher precisions and recalls in all subtasks by significantly reducing the invalid labelling sequences (e.g., \texttt{I-arg} appears right after \texttt{O}). During prediction, instead of tagging each token independently, BLSTM-CRF takes into account the constraints between neighbouring labels, and potentially increases the co-occurrences of key arguments regarding the same event type.

As shown in Table~\ref{tab:1}, the ILP-based post inference considerably improves the overall performance, especially in \textit{event type classification}. With the help of constraint \textbf{C4},  dubious key arguments can be correctly inferred through other key arguments from their context. Compared with BLSTM-CRF, BLSTM-CRF-ILP$_1$ produces an F1 gain of 3.2\% in event type classification, 1.5\% in key argument detection, and 1.2\% in all argument detection.

\noindent \textbf{\emph{Multiple Type Events}:  \mbox{ }} Among all methods, BLSTM-CRF-ILP$_{multi}$ is the only model that can deal with multiple type event mentions. 
The proposed strategy ILP$_{multi}$ helps detect more event mentions for a sentence, contributing to the increase of recalls, and F1 scores with a little drop of precisions.
BLSTM-CRF-ILP$_{multi}$ can correctly identify 132 sentences with multiple type events, with an accuracy of 95.6\%, and for each involved event, our model maintains a high
performance in identifying its arguments, achieving 45.5\%, and 29.1\% in F1 for key argument detection and all argument detection, respectively.

\subsection{Manual Evaluation}\label{manualeve}
To provide a deep investigation about our dataset and models, we randomly sample 150 sentences from the test set. Two annotators are asked
to annotate each sentence following two steps. First, determine if a given sentence is positive or negative, and assign \textbf{all possible} event types to positive ones. Next, label all related arguments and their roles according to the event types for all positive
instances. Two annotators will independently annotate each sentence, and discuss to reach an agreement. The inter-annotator agreement is
87\% for event types and 79\% for arguments.

\begin{figure}[t!]
	\centering
	\includegraphics[width=.47\textwidth]{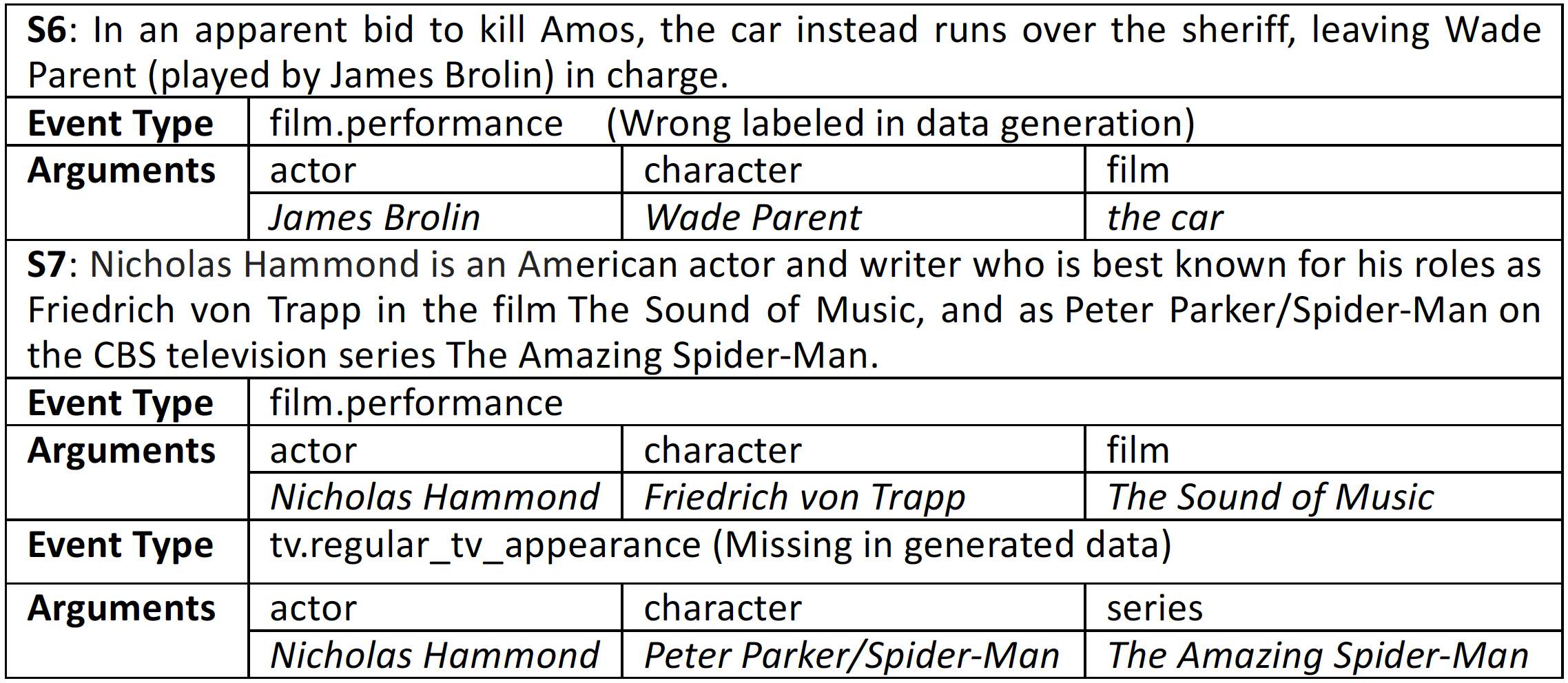}
	\caption{Example outputs of BLSTM-CRF-ILP$_{multi}$.\label{fig:1}}
\end{figure}

By comparing the automatic and manual annotations on the 150 sentences, we find that the main issue for the automatic annotation is that
some automatically labeled sentences do not imply any event while still matching all key properties of certain \CVT entries in Freebase. We
find 16 such instances that are mistakenly labeled as positive. For example in Figure~\ref{fig:1}, although the phrase \underline{\emph{the
car}} in S6 matches a film name, it does not refer to a film. This is because that we currently do not have a strong entity linker to
verify those entities, which we leave for future work. However, during manual investigation, BLSTM-CRF-ILP$_{multi}$ can correctly identify
these 6 instances as negative.

On this manually annotated dataset, we can observe similar trends with Table~\ref{tab:1}, and BLSTM-CRF-ILP$_{multi}$ remains the best performing model, achieving 80.7\%, 56.4\% and 36.3\% in F1 scores for event type, key argument detection and all argument detection, respectively.

Remarkably, our BLSTM-CRF-ILP$_{multi}$ model can find more \CVT instances that are currently not referenced in Freebase. Our model detects two events in S7, while the arguments of the \textit{tv.tv\_appearance} event do not match any existing CVT instances in Freebase, which do not receive any credit during automatic evaluation, but should be populated into Freebase. 
This  suggests that by learning from distant supervision provided by Freebase, our model can be used to populate or update Freebase instances in return.

\noindent \textbf{\emph{On BBC News}:  \mbox{ }}
We further apply our event extractor, trained on FBWiki, to 397 BBC News
articles (2017/04/18 -- 2017/05/18 in Politics, Business and TV sections), and manually examine the extraction results. We find that our model is able to correctly identify 117 events, and 53 events, almost half of which are not covered in the currently used Freebase.

\subsection{Tables as Indirect Supervision}
To investigate the applicability of our approach to other structured knowledge/tables besides Freebase \CVT tables, we automatically build
a new dataset, \textbf{TBWiki}, with the supervision provided by Wikipedia tables, which characterize events about business acquisition,
winning of the Olympics games, and awards winning in entertainment (Table~\ref{tab:6}).

\begin{table}[t]
	\scriptsize
	\centering
	\begin{tabular}{lccccc}
		\toprule
		\textbf{Event type} & \textbf{Entries} & \textbf{Positive} & \textbf{EC} & \textbf{KAD} & \textbf{AAD} \\
		\midrule
		\rowcolor{Gray} Acquisition & 690 & 414 & 87.0\% & 72.0\% & 69.6\% \\
		Olympics & 2503 & 1460 & 77.2\% & 64.5\% & 38.6\% \\
		\rowcolor{Gray} Awards & 3039 & 2217 & 95.0\% & 82.8\% & 58.6\% \\
		\bottomrule
	\end{tabular}
	\caption{Statistics of the TBWiki dataset and the performance (in F1) of our model on TBWiki.
		EC, KAD and AAD denote event type classification, key argument detection and all key argument detection, respectively.
		\label{tab:6}}
\end{table}

We train our BLSTM-CRF-ILP$_{multi}$ on this dataset and evaluate it on 100 manually annotated sentences.
We can see that without extra human annotations, 
our model can learn to extract events from the training data weakly supervised by Wikipedia tables. Given a specific event type, as long as we can acquire tables implying events of such type, it is possible to automatically collect training data from such tables, and learn to extract structured event representations of that type.

\section{Related Work}
Most event extraction works are within the tasks defined by several evaluation frameworks (e.g., MUC~\cite{grishman1996message}, 
ACE~\cite{doddington2004automatic}, ERE~\cite{song2015light} and TAC-KBP~\cite{mitamura2015event}), 
all of which can be considered as a template-filling-based extraction task.
These frameworks focus on limited number of event types, which are designed and annotated by human experts and
hard to generalize to other domains.  
Furthermore, existing extraction systems, which usually adopt a supervised learning paradigm, 
have to rely on those high-quality training data within those frameworks, 
thus hard to move to more domains in practice, regardless of feature-based~\cite{gupta2009predicting,hong2011using,li2013joint} or neural-network-based methods~\cite{chen2015event,nguyen2016joint}.

Besides the works focusing on small human-labeled corpora, 
Huang et al. \shortcite{huang2016liberal} and Chen et al.\shortcite{chen2017automatically} leverage various linguistic resources (e.g., FrameNet, VerbNet, etc.) to automatically collect trigger annotations 
with more training instances to improve existing event extractors.
In contrast, we propose to exploit various structured knowledge bases to automatically discover 
types of events as well as their corresponding argument settings, without expert annotations, and further automatically
construct training data, with the help of \DS.

Distant supervision~(\DS) has been widely used in binary relation extraction, where the key assumption is that 
sentences containing both the subject and object of a $<$$subj$, $rel$, $obj$$>$ triple can be seen as its support, and further
used to train a classifier to identify the relation $rel$. However,  this assumption does not fit to our event extraction scenario, 
where an event usually involves several arguments and it is hard to collect enough training sentences with all arguments appearing in, as indicated by the low coverage of \textit{ALL}. We therefore investigate different generation strategies for event extraction within the \texttt{DS} paradigm and propose to utilize time and syntactic clues to refine the \DS assumption for better data quality. 
We further relieve the reliance on explicit trigger annotations required by previous event extractors, and define a novel event extraction paradigm with key arguments to characterize an event type. 

\section{Conclusions}
This paper has presented a novel, fast approach to automatically construct training data for event extraction with little human
involvement, which in turn allows effective event extraction modeling. To generate training data, our approach first extracts, from
existing structured knowledge bases, which of the arguments best describe an event; then, it uses the key arguments to automatically infer the occurrence of an event without explicit trigger identification. To perform event extraction, we develop a novel architecture based on neural networks and post inference, which does not require explicit trigger information. We apply our approach to label Wikipedia articles using knowledge extracted from various knowledge bases. We show that the quality of the automatically generated training data is comparable to those that were manually labeled by human experts. We demonstrate that this large volume of high-quality training data, combining with our novel event extraction architecture, not only leads to a highly effective event extractor, but also enables multiple typed event detection.

\section{Acknowledgments}
This work is supported by National High Technology R\&D Program of China (Grant No.2015AA015403), Natural Science Foundation of China (Grant No. 61672057, 61672058, 71403257); the UK Engineering and Physical Sciences Research Council under grants EP/M01567X/1 (SANDeRs) and EP/M015793/1 (DIVIDEND); and the Royal Society International Collaboration Grant (IE161012). For any correspondence, please contact Yansong Feng. 

\bibliography{aaai}
\bibliographystyle{aaai}
\end{document}